\documentclass[10pt,twocolumn,letterpaper]{article}

\usepackage{cvpr}              

\usepackage[dvipsnames]{xcolor}

\usepackage{multirow}

\usepackage{pifont}
\usepackage{color}
\usepackage{newtxmath}
\usepackage{amsmath, mathtools}
\definecolor{cvprblue}{rgb}{0.21,0.49,0.74}
\usepackage[pagebackref,breaklinks,colorlinks,citecolor=cvprblue]{hyperref}

\usepackage{graphics}           
\usepackage{times}              
\usepackage{amsmath}            
\usepackage{amssymb}            
\usepackage{graphicx}
\usepackage{algorithm}
\usepackage[noend]{algpseudocode}
\usepackage{booktabs}
\usepackage{color}
\definecolor{instructioncolor}{rgb}{.5,.5,.5}

\usepackage[font=small]{caption}

\def\secref#1{Sec.~\ref{#1}}
\def\figref#1{Fig.~\ref{#1}}
\def\tabref#1{Tab.~\ref{#1}}
\def\eqref#1{Eq.~(\ref{#1})}

\def\etalcite#1{\etal~\cite{#1}}

\usepackage{array}
\newcolumntype{L}[1]{>{\raggedright\let\newline\\\arraybackslash\hspace{0pt}}m{#1}}
\newcolumntype{C}[1]{>{\centering\let\newline\\\arraybackslash\hspace{0pt}}m{#1}}
\newcolumntype{R}[1]{>{\raggedleft\let\newline\\\arraybackslash\hspace{0pt}}m{#1}}

\def\argmax{\mathop{\rm argmax}}

\newcommand{\norm}[1]{\lVert#1\lVert}

\renewcommand{\b}[1]{\mbox{\boldmath$#1$}}

\newcommand{\cmark}{\ding{51}}%
\newcommand{\xmark}{\ding{55}}%
\def\paperID{1784} 
\def\confName{CVPR}
\def\confYear{2024}

\title{Open-World Semantic Segmentation Including Class Similarity}

\author{
Matteo Sodano$^1$
\quad
Federico Magistri$^1$
\quad
Lucas Nunes$^1$
\quad
Jens Behley$^1$
\quad
Cyrill Stachniss$^{1,2}$ \\
\quad
{\small $^1$Center for Robotics, University of Bonn \qquad \qquad $^2$Lamarr Institute for Machine Learning and Artificial Intelligence}\\
{\tt\small \{firstname.lastname\}@igg.uni-bonn.de}
\vspace{-1em}
}

\begin{document}

\twocolumn[{%
\renewcommand\twocolumn[1][]{#1}%

\maketitle
\thispagestyle{empty}
\pagestyle{empty}

}]

\begin{abstract}
    Interpreting camera data is key for autonomously acting systems, such as autonomous vehicles. Vision systems that operate in real-world environments must be able to understand their surroundings and need the ability to deal with novel situations. This paper tackles open-world semantic segmentation, i.e., the variant of interpreting image data in which objects occur that have not been seen during training. We propose a novel approach that performs accurate closed-world semantic segmentation and, at the same time, can identify new categories without requiring any additional training data. Our approach additionally provides a similarity measure for every newly discovered class in an image to a known category, which can be useful information in downstream tasks such as planning or mapping. Through extensive experiments, we show that our model achieves state-of-the-art results on classes known from training data as well as for anomaly segmentation and can distinguish between different unknown classes. 
\vspace{-10pt}
\end{abstract}
    
\section{Introduction}
\label{sec:intro}

\begin{figure}[ht!]
  \centering
  \includegraphics[width=0.83\linewidth]{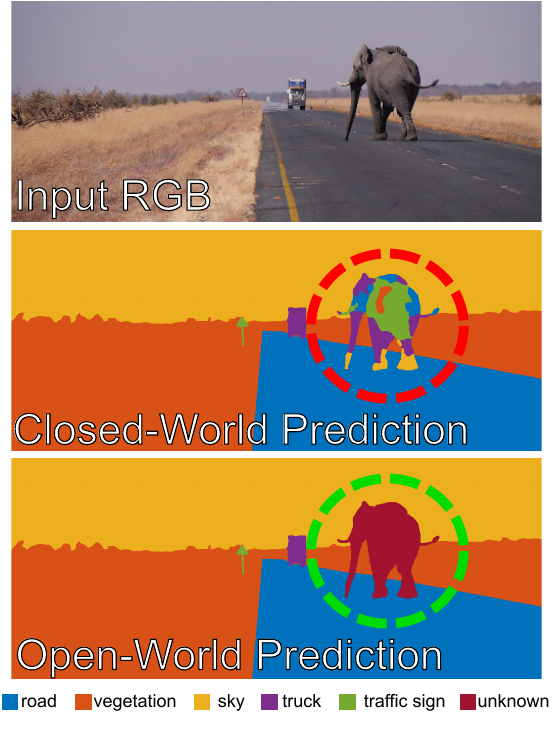}
   \vspace{-8pt}
  \caption{Given an image containing a previously-unseen object (top), closed-world methods for semantic segmentation classify the pixels belonging to that object as one of the known classes (center, red circle). Our goal is to segment the unknown object and identify it as a semantic class different to the previously-known ones (bottom, green circle). }
  \label{fig:motivation}
  \vspace{-16pt}
\end{figure}


Autonomous systems need to understand their surroundings to operate robustly. To this end, semantic scene understanding based on sensor data is key and numerous variants exist, such as object detection~\cite{girshick2015iccv,ren2015nips}, semantic and instance segmentation~\cite{long2015cvpr-fcnf, marks2023ral, xie2021neurips}, and panoptic segmentation~\cite{kirillov2019cvpr-ps, kirillov2019cvpr}. Over the last decade, we witnessed tremendous progress in scene interpretation for autonomous vehicles using machine learning. A central challenge for most learning-based systems is scenes in which novel and previously unseen objects occur. Such \textit{open-world} settings, i.e., the fact that not everything can be covered in the training data, have to be considered when building vision systems for human-centered environments and real-world settings. For example, autonomous cars in cities will eventually experience situations or objects they have not seen before. They should be able to identify them, for example, to change into a more conservative mode of operation.


Today, high-quality datasets such as Cityscapes~\cite{cordts2016cvpr} or MS COCO~\cite{lin2014eccv} allow deep learning methods to achieve outstanding performance in closed-world scene understanding tasks. A prominent task is semantic segmentation~\cite{girshick2014cvpr}, which aims to assign a semantic category to each pixel in an image. Systems operating under the \textit{closed-world} assumption~\cite{russell2010book} typically cannot correctly recognize an object that belongs to none of the known categories. Often, they tend to be overconfident and assign such an object to one of the known classes. We believe that for applications targeting reliability and robustness under varying conditions, the closed-world assumption has to be relaxed, and we need to move towards open-world setups. Additionally, a measure of class similarity can help downstream tasks. For example, predicting that an area of the image is unknown but similar to the class car or another type of moving vehicle can be used in planning or tracking to estimate the motion of the object, or in mapping to discard that class from the map.


This paper investigates the problem of \textit{open-world} semantic segmentation. Given an image at test time, we aim to have a model that is able to detect any pixel that belongs to a category that was unseen at training time and is also able to distinguish between different new categories. The first problem is called \textit{anomaly segmentation}~\cite{chan2021neurips} and aims to achieve a binary segmentation between known and unknown. The second problem, called \textit{novel class discovery}~\cite{han2019cvpr}, aims to obtain a pixel-wise classification of novel samples into different classes starting from the knowledge of previously seen, labeled samples. We aim to investigate how to solve both tasks jointly in a neural network setting. We extend best-practice approaches for anomaly detection for classification tasks~\cite{bendale2016cvpr,dhamija2018neurips,yao2023cvpr} and provide compelling results for both, anomaly segmentation and novel class discovery. See~\figref{fig:motivation} for an example of the targeted output.


The main contribution of this paper is a novel approach for open-world segmentation based on an encoder-decoder convolutional neural network (CNN). We propose a new architecture that simultaneously performs accurate closed-world semantic segmentation while constraining all known classes towards their learned feature descriptor, thanks to a loss function we introduce. We combine operations on the feature space with binary anomaly segmentation that allows us to distinguish between different novel classes and provide a measure of similarity for every newly discovered class to a known category. 
We implemented and thoroughly tested our approach. 
In sum, our contributions are the following:
\begin{itemize}
  \item A fully-convolutional neural network that achieves state-of-the-art performance on anomaly segmentation while providing compelling closed-world performance.
  \item A loss function that allows us to distinguish among different novel classes, and to provide a similarity score for each novel class to the known categories.
  \item Extensive experiments on multiple datasets, including the public benchmark SegmentMeIfYouCan, where we rank first in three out of five metrics.
\end{itemize}
\section{Related Work}
\label{sec:related}

Semantic segmentation under closed-world settings achieved outstanding performances in different domains, such as autonomous driving~\cite{borse2021cvpr,milioto2019icra,milioto2019icra-fiass,wang2023cvpr}, indoor navigation~\cite{hu2021cvpr,kundu2020eccv,sodano2023icra}, or agricultural robotics~\cite{ciarfuglia2023compag,milioto2018icra,roggiolani2022icra,weyler2023arxiv}. However, the closed-world assumption should be relaxed when developing systems for navigating in the wild. In such cases, we need to move towards open-world setups.

\textbf{Anomaly Detection and Classification}.
The open-world setting was initially explored for classification, where anomalous samples had to be recognized and discarded. This problem was tackled in different ways in the literature. Simple strategies such as thresholding the softmax activations~\cite{cardoso2017ml,hendrycks2017iclr}, using a background class for tackling unknown samples~\cite{blum2021ijcv,mor2018wacv,vaze2021iclr}, and using model ensembles~\cite{lakshminarayanan2017nips,vyas2018eccv} represent a solid starting point in theory. In practice, however, closed-world predictions tend to be overconfident by showing a peak in the softmax even for unknown samples~\cite{nguyen2015cvpr,tommasi2017dacva}. Additionally, it is impossible to train with all possible examples of unknown objects. To deal with this, modifications to the softmax layer have been proposed~\cite{bendale2016cvpr}. Other approaches rely on maximizing the entropy~\cite{dhamija2018neurips} or on energy scores~\cite{liu2020neurips}, which are supposedly less susceptible to the aforementioned overconfidence issue. Even though these approaches can be easily adapted to the segmentation problem, they are limited in the sense that they rely on the output of the CNN to be ``uncertainty-aware'' to some extent, in order to be able to modify the scores, consider the output entropies, or similar. 

In contrast, we operate on the feature space of the semantic segmentation to not only classify pixels correctly but also match their feature to a unique class descriptor, in order to use the distance from it as a measure of ``unknown-ness''.

\begin{figure*}[t]
  \centering
  \includegraphics[width=0.9\linewidth]{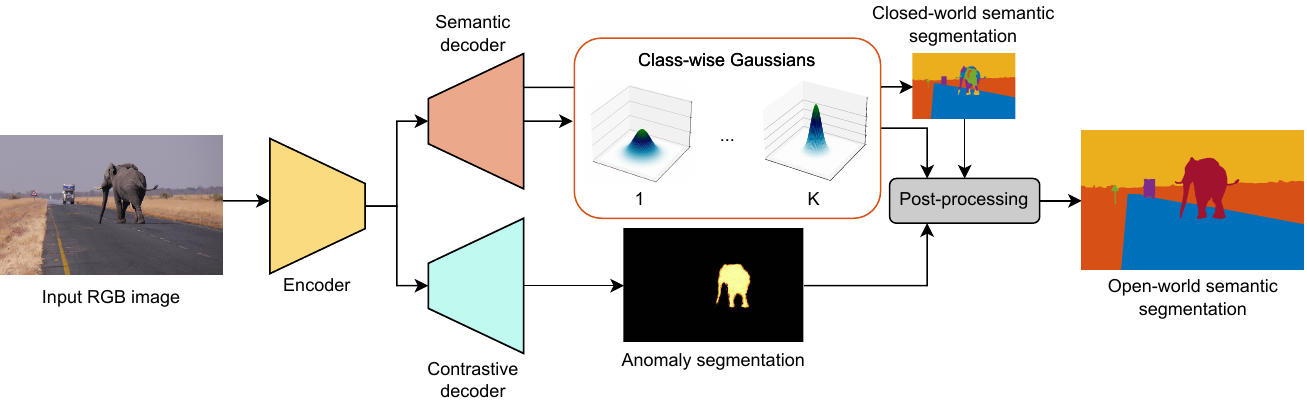}
  \caption{Given an RGB image as input, our network processes it by means of an encoder and two decoders. The semantic decoder produces a closed-world semantic segmentation and a Gaussian model for each known category. The class Gaussian models are built from a learned class descriptor (mean) and the variance of all predictions from it. A 3D example is shown in the image. The contrastive decoder provides an anomaly segmentation output. A post-processing phase finally achieves open-world semantic segmentation.}
  \label{fig:architecture}
  \vspace{-1em}
\end{figure*}
\textbf{Open-World Segmentation}.
Open-world or anomaly segmentation extends the anomaly detection task to a pixel-wise nature, by trying to predict whether each individual pixel in an image is an anomaly or not. Some methods rely on the estimation of the uncertainty on the prediction with Bayesian deep learning~\cite{gal2016icml,lakshminarayanan2017nips,sapkota2022cvpr}, or on the gradient itself~\cite{liu2018cvpr-ffpf,maag2023arxiv}. Other works use an additional dataset for out-of-distribution samples, in order to help the CNN recognize categories that do not belong to the standard training set~\cite{bevandic2019pr,chan2021cvpr}. Recently, generative models have also been used for such task, since in the reconstruction phase they will accurately resynthesize only the known areas, while unknown objects will suffer from a lower reconstruction quality, and can be recognized by looking at the most dissimilar areas between the input and the output~\cite{kong2021cvpr,lis2020arxiv,zhao2023cvpr-oauc}. Due to the limitation of available training data, many unsupervised approaches use synthetic anomaly data and train an anomaly detector which is either distance-based~\cite{liu2023cvpr-dad,roth2022cvpr,tsai2022wacv} or reconstruction-based~\cite{bergmann2020cvpr,zavrtanik2021iccv,zhang2023cvpr-dsgd}, with the latter sharing the same concept as the generative models mentioned above. Vision-language models based on CLIP~\cite{radford2021icml,rao2022cvpr-dldp,zhong2022cvpr-rrlp} are also gaining interest in the context of anomaly segmentation~\cite{jeong2023cvpr-wzac}. Lately, a lot of research interest is also going in the direction of anomaly segmentation in video streams because of its application in intelligent surveillance systems~\cite{sun2023cvpr-hscf,yang2023cvpr,zhang2023cvpr-ecau}.

Differently from these approaches, we do not require additional data for training and do not rely on uncertainty estimation or generative models. In contrast, by operating on the feature space of the semantic segmentation task, we can define a distinct region for known and unknown classes. Furthermore, we leverage the feature descriptors of the known classes to recognize different unknown classes and find the most similar known class. 

\section{Our Approach}
\label{sec:main}

In this work, we tackle the problem of open-world semantic segmentation. In addition to handling known classes, we are particularly interested in segmenting all anomalous areas in an image, where previously unseen objects appear, and in differentiating between potentially multiple novel classes. We propose an approach (see~\figref{fig:architecture}) based on a convolutional neural network with one encoder and two decoders. The first decoder tackles semantic segmentation and operates on the feature space so that, for each class, features of pixels belonging to the same class are pushed together. The mean and variance of each individual class descriptor are stored representing Gaussian distributions that describe known classes. The second decoder performs binary anomaly segmentation. Results are finally merged to obtain open-world semantic segmentation, i.e. anomaly segmentation and novel class discovery.

\subsection{General Network Architecture}
\label{subsec:arch}
Our network for open-world semantic segmentation is composed of one encoder and two decoders. We use a ResNet34~\cite{he2016cvpr} encoder, where the basic ResNet block is replaced with the NonBottleneck-1D block~\cite{romera2018tits}, which allows a more lightweight architecture since all $3 \times 3$ convolutions are replaced by a sequence of $3 \times 1$ and $1 \times 3$ convolutions with a ReLU in between. For open-world segmentation, contextual information is valuable. Therefore, we expand the limited receptive field of ResNet by incorporating contextual information using a pyramid pooling module~\cite{zhao2017cvpr-pspn} after the encoding part. The features produced will be fed to two separate decoders, that share the same structural properties. In order to preserve the lightweight nature of the network, we use three SwiftNet modules~\cite{orsic2019cvpr} where we incorporate NonBottleneck-1D blocks, and two final upsampling modules based on nearest-neighbor and depth-wise convolutions, which reduce the computational load. 
We use encoder-decoder skip connections after each downsampling stage of the encoder to directly propagate more fine-grained features to the decoder.

\subsection{Approach for Open-World Segmentation}
\label{subsec:approach_ow}
Our approach for open-world segmentation builds upon the structure of the CNN we developed, and it exploits the double-decoder architecture for providing accurate segmentation of unknown regions. The first decoder, which we call ``semantic decoder'' in the following, targets semantic segmentation. We additionally manipulate the feature space to create a unique distinct descriptor for each known class. Our goal is to obtain a correct semantic segmentation for the known classes, but also produce pre-softmax features that are similar to the descriptor for each pixel of a certain class. In this way, we aim to detect as unknown classes all pixels whose feature vectors are substantially different from the descriptor of the class they have been assigned to. 
The second decoder, which we call ``contrastive decoder'' in the following, leverages the contrastive loss~\cite{chen2020icml} and objectosphere loss~\cite{dhamija2018neurips} together, to place all features of known classes on the surface of a hypersphere while pushing the ones of unknown classes towards its center. In this way, the second decoder provides an anomaly segmentation, where the anomalous regions correspond to previously unseen classes. The two results are finally merged using an automated post-processing operation to obtain the final open-world segmentation.

In the following, we call $\Omega = \{(1,\,1),\dots,(H,\,W)\}$ the set of pixels in the image, $Y \in \{1,\dots,K\}^{H \times W}$ the ground truth mask, and $\hat{Y} \in \{1,\dots,K\}^{H \times W}$ the predicted mask, where $H$ and $W$ are the dimensions of the input image. Additionally, we denote with \mbox{$\Omega_k = \{ p \in \Omega \mid Y_p = k \}$} the set of pixels whose ground truth label is $k$, and with \mbox{$\hat{\Omega}_k = \{p \in \Omega \mid \hat{Y}_p = Y_p \}$} the set of pixels that are true positives for class $k$, \ie, the set of pixels whose ground truth label and predicted label are $k$. Finally, the square of a vector refers to the element-wise operation (Hadamard product):
\begin{equation} 
  \b{v}^2 = \begin{bmatrix} v_1^2, \, \dots, \, v_n^2 \end{bmatrix}^{\top}
\end{equation}

\textbf{Semantic Decoder}. The aim of semantic segmentation is to predict a categorical distribution over $K$ classes for all pixels in an image. We follow best practice and optimize it with the weighted cross-entropy loss
\begin{equation}
  \mathcal{L}_{\mathrm{sem}} = -\frac{1}{|\Omega|} \sum_{p \in \Omega} \omega_k \b{t}_p^\top \log\big( \sigma (\b{f}_p) \big),
  \label{eq:loss_semantic}
\end{equation}
where $\omega_k$ is a class-wise weight computed via the inverse frequency of each class in the dataset, $\b{t} \in \mathbb{R}^{H \times W \times K}$ is a one-hot encoded pixel-wise ground truth label, $\b{t}_p \in \mathbb{R}^K$ is a one-hot encoded pixel-wise ground truth label at pixel \mbox{$p \in \Omega$}, $\sigma$ indicates the softmax operation, and $\b{f}_p$ denotes the pre-softmax feature predicted for pixel $p$.

As mentioned above, we do not only want to perform standard semantic segmentation but also build a class descriptor to bring all pixels belonging to a certain class towards a certain region in the feature space. To achieve this, we accumulate the pre-softmax features, also called activation vectors, of all true positives for each class, where a true positive is a pixel that is correctly segmented. With this, we can store a running average class prototype, or mean activation vector, $\b{\mu}_k \in \mathbb{R}^K$ for each class $k \in \{ 1, \dots, K \}$:
\begin{equation}
  \b{\mu}_k = \dfrac{1}{|\hat{\Omega}_k|} \sum_{p \in \hat{\Omega}_k} \b{f}_p.
\end{equation}

We also iteratively compute the per-class variance \mbox{$\b{\sigma}_k^2 \in \mathbb{R}^K$} via sum of squares, as
\begin{figure}[t]
  \centering
  \includegraphics[width=0.9\linewidth]{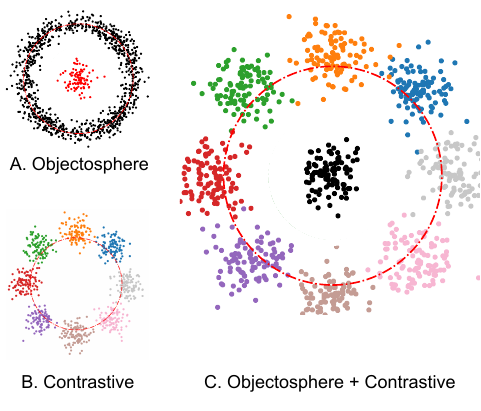}
  \caption{2D visualization of the expected output of the contrastive decoder. The behavior of the objectosphere loss is shown in A, where all points coming from known classes (black) lie around the red (outer) circle of radius $\xi$, see~\eqref{eq:loss_objectosphere}, and the points from unknown classes lie around the origin. The contrastive loss is shown in B, where features lie on the unit circle. Together, they lead to a behavior similar to the one depicted in C.}
  \label{fig:contrastive_dec}
  \vspace{-10pt}
\end{figure}
\begin{equation}
  \b{\sigma}_k^2 = \dfrac{1}{|\hat{\Omega}_k|}  \sum_{p \in \hat{\Omega}_k} \, \big( \b{f}_{p} - \b{\mu}_k \big)^2.
  \label{eq:variance}
\end{equation}

At the beginning of epoch $e$, we have the means $\b{\mu}_k^{e-1}$ and variances $\b{\sigma}_k^{e-1}$ accumulated in the previous epoch.
At epoch $e$, we can steer the semantic segmentation to predict, for each pixel with ground truth class $k$, a feature vector equal to $\b{\mu}_k^{e-1}$. For this, we introduce a feature loss function
\begin{equation}
  \mathcal{L}_{\mathrm{feat}} = \frac{1}{|\Omega|} \sum_{k=1}^{K} \sum_{p \in \,\Omega_k} \frac{\norm{\b{f}_p - \b{\mu}_{k}^{e - 1}}}{\b{\sigma}_{k}^{e-1}}.
  \label{eq:loss_mavs}
\end{equation}

This loss function is not active during the first epoch since there is no accumulated mean yet. Thus, we perform standard semantic segmentation during the first epoch.

The semantic decoder is thus optimized with a weighted sum of the loss functions introduced above
\begin{equation}
  \mathcal{L}_{\mathrm{sdec}} = w_1 \, \mathcal{L}_{\mathrm{sem}} + w_2 \, \mathcal{L}_{\mathrm{feat}}.
  \label{eq:loss_decoder1}
\end{equation}

\textbf{Contrastive Decoder}. The contrastive decoder explicitly aims for anomaly segmentation. Given an image of dimensions $H \times W$, where known and unknown classes are present, the goal of the contrastive decoder is to provide the basis for a binary prediction where $0$ corresponds to known classes and $1$ to unknown classes. We achieve this by means of a combination between the contrastive loss~\cite{chen2020icml} and the objectosphere loss~\cite{dhamija2018neurips}. First, we compute the mean feature representation $\overline{\b{f}}_k$ for class $k$ in the current image as

\begin{equation}
	\overline{\b{f}}_k = \frac{1}{|\Omega_k|} \sum_{p \in \Omega_k} \b{f}_p^{\, d},
  \label{eq:loss_mean_feat}
\end{equation}
where $\b{f}_p^d$ is the feature predicted at pixel $p$ from the contrastive decoder (the equivalent of $\b{f}_p$ for the semantic one). 
Then, we compute the contrastive loss $\mathcal{L}_{\mathrm{cont}}$ such that $\overline{\b{f}}_k$ approximates the normalized mean representation $\bar{\b{\mu}}^{e-1}_k$ of the corresponding class in the previous epoch $\b{\mu}^{e - 1}_k$ and gets dissimilar from the other classes mean representation:

\begin{equation}
	\mathcal{L}_{\mathrm{cont}} = - \sum_{k=1}^{K} \log \frac{\exp{({\overline{\b{f}}_k}^\top \bar{\b{\mu}}_{k}^{e-1} / \tau})}{\sum_{i=1}^K{\exp{({{\overline{\b{f}}_k}^\top \bar{\b{\mu}}_i^{e-1} / \tau})}}},
  \label{eq:loss_contrastive}
\end{equation}

\noindent where $\tau$ is a temperature parameter. This way, the loss aims to make the features from the same class consistent with its running mean representation $\b{\mu}_k^{e-1}$, while scattering all $K$ classes around the unit hypersphere. 

At the same time, we use the objectosphere loss $\mathcal{L}_{\mathrm{obj}}$ over each pixel $p \in \Omega$ given by

\begin{equation}
  \mathcal{L}_{\mathrm{obj}} = \left\{ \begin{array}{cl}
    \max \big( \xi - \norm{\b{f}_{p}}^2, \, 0 \big) &, \mathrm{if}  \, p \in \mathcal{D}_{k} \\ 
    \norm{\b{f}_{p}}^2  &, \mathrm{otherwise} \end{array}\right.,
  \label{eq:loss_objectosphere}
\end{equation}
where $\mathcal{D}_{k}$ is the set of pixels belonging to known classes. The remaining pixels, at training time, reduce to the unlabeled (void) areas of the image. This aims to make the norm of the feature vector $\norm{\b{f}_p}$ of pixels belonging to known classes $\mathcal{D}_k$ bigger than a threshold~$\xi$, while the norm of the features of pixels belonging to unknown classes $\mathcal{D}_u$ gets reduced to~$0$. These two loss functions~$\mathcal{L}_{\mathrm{cont}}$ and~$\mathcal{L}_{\mathrm{obj}}$ together allows us to optimize towards a situation where the feature vectors of known classes are distributed along the surface of the $K$-dimensional hypersphere of radius $\xi$, while the feature vectors of unknown classes gets squashed to $0$. A 2D example of the expected behavior is shown in~\figref{fig:contrastive_dec}.

The contrastive decoder is optimized with a weighted sum of the two losses given by
\begin{equation}
  \mathcal{L}_{\mathrm{cdec}} = w_3 \, \mathcal{L}_{\mathrm{cont}} + w_4 \, \mathcal{L}_{\mathrm{obj}}.
  \label{eq:loss_decoder2}
\end{equation}
 
\begin{figure*}[t]
  \centering
  \includegraphics[width=0.9\linewidth]{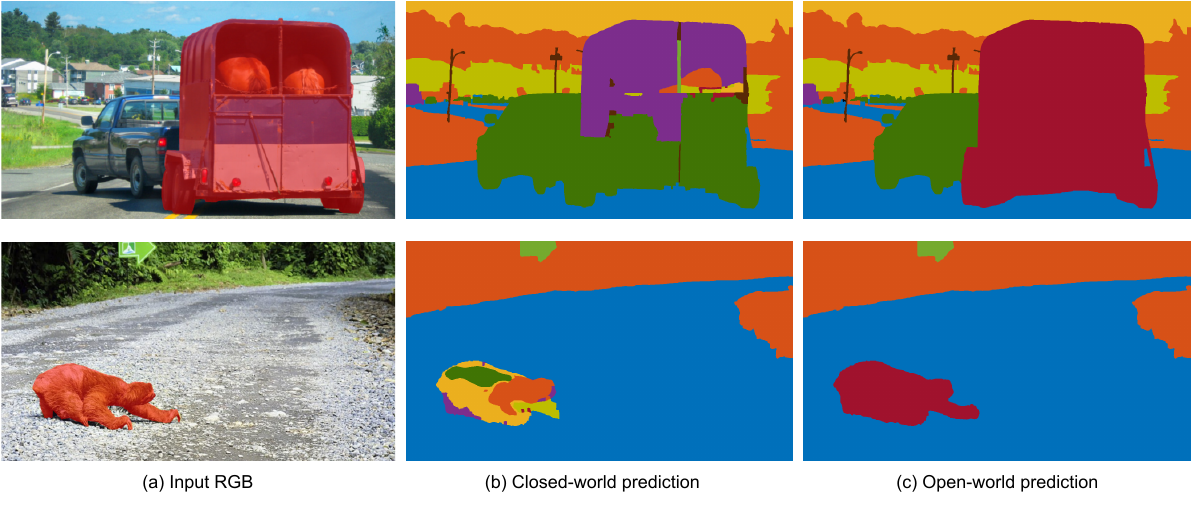}
  \vspace{-0.5cm}
  \caption{Results from the validation set of SegmentMeIfYouCan. We show the input RGB overlayed with the ground truth unknown mask (a), the prediction of our closed-world model (b), and the prediction of our approach for open-world segmentation (c). In the open-world prediction, the unknown class is shown in red.}
  \label{fig:results_smiyc}
\end{figure*}
\textbf{Post-Processing for Anomaly Segmentation}. To obtain the open-world predictions at test time, we fuse the outputs of the two decoders. 
The semantic encoder provides a standard closed-world semantic segmentation but, thanks to the loss function that operates directly on the feature space that we introduced, we can obtain an open-world segmentation. In fact, we computed mean $\b{\mu}_k \in \mathbb{R}^K$ and variance~$\b{\sigma}_k^2 \in \mathbb{R}^K$ of each class, meaning that, for each class, we can easily build a multi-variate normal distribution $\mathcal{N}\big( \b{\mu}_k, \, \b{\Sigma}_k \big)$, where $\b{\mu}_k$ is the mean, and $\b{\Sigma}_k = \mathrm{diag}(\b{\sigma}_k^2)$ is the covariance matrix, which reduces to the diagonalization of the variance $\b{\sigma}_k^2$ under the assumption that all classes are independent. After building the Gaussian model of each class in the dataset, given a pixel~$p$ whose predicted feature~$\b{f}_p$ would correspond to class~$k, \forall k$, we compute a fitting score by means of the squared exponential kernel
\begin{equation}
  s_k(\b{f}_p) = \exp{\Bigg(- \frac{1}{2} (\b{f}_p - \b{\mu}_k)^\top \b{\Sigma}_k^{-1} (\b{f}_p - \b{\mu}_k) \Bigg)}. 
  \label{eq:gaussian_pdf}
\end{equation}

Then, for each pixel, we take the highest score
\begin{equation}
  s(p) = \max_{k} s_k(\b{f}_p), 
  \label{eq:max_score}
\end{equation}
and, if it is low, then the pixel of interest is in the tail of the Gaussian, and is considered as a novel class, leading to an open-world prediction $\mathcal{U}_{\mathrm{sem}}$ of the semantic decoder. We can obtain a pixel-wise score $s_{\mathrm{unk}, \, p}^{\mathrm{sem}}$ for being unknown $
  s_{\mathrm{unk}, \, p}^{\mathrm{sem}} = 1 - s(p).
$

The contrastive decoder leads to a second open-world prediction $\mathcal{U}_{\mathrm{cont}}$ by considering as unknown all pixels whose feature norm is below a certain threshold. In particular, we can obtain a pixel-wise score $s_{\mathrm{unk}, \, p}^{\mathrm{cont}}$ for being unknown
\begin{equation}
  s_{\mathrm{unk}, \, p}^{\mathrm{cont}} = \max \Bigg( 0, \, \Bigg(1 - \frac{\norm{\b{f}_p}^2}{\xi} \Bigg) \Bigg),
  \label{eq:score_contrastive}
\end{equation}
where $\b{f}_p$ is the predicted feature at pixel $p$. This score is $1$ when the norm of the feature vector is 0, and $0$ when the norm is bigger than $\xi$, as described in \eqref{eq:loss_objectosphere}.

Finally, we fuse the two predictions to obtain a cumulative pixel-wise score for being unknown as
\begin{equation}
  s_{\mathrm{unk}, \, p} = \frac{1}{2} \, \Big( s_{\mathrm{unk}, \, p}^{\mathrm{sem}} + s_{\mathrm{unk}, \, p}^{\mathrm{cont}} \Big).
  \label{eq:score_final}
\end{equation}

If $s_{\mathrm{unk}, \, p}$ is above a threshold $\delta$, the pixel is considered belonging to an unknown class.

\textbf{Post-Processing for Open-World Semantic Segmentation}. When a pixel is considered unknown, we need to store its activation vector and decide whether it belongs to an already-discovered class or a new one. Given the set of mean activation vectors for $G$ unknown classes discovered so far $\mathcal{F} = \{\b{f}_u^1, \dots, \,  \b{f}_u^G\}$, we take the vector $\b{f}_u^g$ that minimizes the distance from the querying vector. If the distance between $\b{f}_u^g$ and $\b{f}_p$ is below a threshold $\eta$, then the pixel belongs to this class, and the mean activation vector gets updated, otherwise it creates a new unknown class $\b{f}_u^{g+1}$. This allows us to have a virtually unlimited number of novel classes.

\subsection{Class Similarity}
\label{subsec:approach_cs}
As a byproduct of the open-world segmentation, our method can also predict the most similar known category for each unknown sample. As explained in~\secref{subsec:approach_ow}, it does not suffice for a feature vector to have the highest activation in the $k$-th spot for being matched to class $k$. A sample can have the highest activation for a certain class $k$ but its score computed with~\eqref{eq:gaussian_pdf} is higher for another class $\tilde{k} \neq k$, meaning that the sample is more inside the area of influence of class~$\tilde{k}$ despite having a higher activation on class $k$. As the most similar class, we propose to choose the one that provides the highest score given by $
  \tilde{k} = \argmax_{k} s_k (\b{f}_p).
$

\begin{table*}[!htb]
  \caption{\textbf{Left}. Comparison between closed-world and open-world model on the known classes of the training datasets. Our OW approach does not harm closed-world semantic segmentation. \textbf{Right}. Results from the public leaderboard of the SegmentMeIfYouCan benchmark. We separate methods that use external data, i.e. out of distribution (OoD) data with semantic labels different from the ones in Cityscapes~\cite{chan2021neurips}, during training. Our approach ranked overall top 1 for FPR95, PPV and mean F1, and top 6 for AUPR and sIoU (fourth and sixth, respectively) on January 31st, 2024. }
  \begin{subtable}[t]{.30\linewidth}
    \centering
   \resizebox{0.99\linewidth}{!}{
   \begin{tabular}{ccc} 
     \toprule
     \multirow{3}{*}{\textbf{Approach}} & \multicolumn{2}{c}{\normalsize{\textbf{mIoU}} [\%] $\uparrow$} \\
     \cmidrule(lr){2-3}
     & CityScapes & BDDAnomaly \\
     \midrule
     CW & 71.1 & 64.1 \\
     OW & 70.8 & 62.8 \\
     \bottomrule
   \end{tabular}
   }
  \end{subtable}
      \begin{subtable}[t]{.69\linewidth}
        \centering
         \resizebox{1\linewidth}{!}{
           \begin{tabular}{ccccccc} 
            \toprule
            \multirow{3}{*}{\textbf{Approach}} & \multirow{3}{*}{\textbf{OoD}} & \multicolumn{2}{c}{\normalsize{\textbf{Pixel-Level}}} & \multicolumn{3}{c}{\normalsize{\textbf{Component-Level}}} \\
            \cmidrule(lr){3-4}\cmidrule(lr){5-7}
            && AUPR [\%] $\uparrow$ & FPR95 [\%] $\downarrow$ & sIoU gt [\%] $\uparrow$ & PPV [\%] $\uparrow$ & mean F1 [\%] $\uparrow$ \\
            \midrule
            DenseHybrid~\cite{grcic2022eccv} & \cmark & 78.0 & 9.8 & 54.2 & 24.1 & 31.1 \\
            RbA~\cite{nayal2022arxiv} & \cmark & \textbf{94.5} & 4.6 & \textbf{64.9} & 47.5 & 51.9 \\
            \midrule
            Maskomaly~\cite{ackermann2023arxiv} & \xmark & 93.4 & 6.9 & 55.4 & 51.2 & 49.9 \\
            RbA~\cite{nayal2022arxiv} & \xmark & 86.1 & 15.9 & 56.3 & 41.4 & 42.0 \\
            ContMAV (ours) & \xmark & 90.2 & \textbf{3.8} & 54.5 & \textbf{61.9} & \textbf{63.6} \\
          \bottomrule    
        \end{tabular}
         }
      \end{subtable}
      \label{tab:multitab} 
      \vspace{-1em}
\end{table*}
\section{Experimental Evaluation}
\label{sec:exp}

%
The main focus of this work is an approach for open-world semantic segmentation that also provides a measure of class similarity.
We present experiments to show the capabilities of our method. The results of our experiments support our claims, which are:
(i)~our model achieves state-of-the-art results for anomaly segmentation while performing competitively on the known classes, (ii)~our approach can distinguish between different unknown classes, and (iii)~our approach can provide a similarity score for each novel class to the known ones.

\subsection{Experimental Setup}

We use two datasets for validating our method: SegmentMeIfYouCan~\cite{chan2021neurips} and \mbox{BDDAnomaly}~\cite{hendrycks2022icml}. Since ground truths are available for the test set of BDDAnomaly, we use it for ablation studies and experiments on class similarity. 

We evaluate our methods with the metrics proposed in the SegmentMeIfYouCan public benchmark for pixel-level performance: area under the precision-recall curve (AUPR) and the false positive rate at a true positive rate of $95 \%$ (FPR95). For SegmentMeIfYouCan, we report also component-level metrics provided by the benchmark.
As explained, our approach is not limited to anomaly segmentation, but performs open-world semantic segmentation. Thus, we also report the mean intersection-over-union (mIoU) on the known classes, to show that our open-world segmentation approach does not underperform on the known classes when compared to the closed-world equivalent (see~\tabref{tab:multitab}, left). Finally, we report the mIoU between the newly-discovered classes and their respective highest-overlapping ground truth class to be discovered.

In all tables, we call our method ``ContMAV'', where ``Cont'' indicates the contrastive decoder and ``MAV'' the mean activation vector of the semantic decoder.

\textbf{Training details and parameters}. In all experiments, we use the one-cycle learning rate policy~\cite{smith2019aiml} with an initial learning rate of 0.004. We perform random scale, crop, and flip data augmentations, and optimize with Adam~\cite{kingma2015iclr} for 500 epochs with batch size 8. We set \mbox{$\xi = 1$}, \mbox{$\delta = 0.6$}, \mbox{$\tau = 0.1$}, \mbox{$\eta = 0.5$}, and loss weights \mbox{$w_1 = 0.9$}, \mbox{$w_2 = 0.1$}, \mbox{$w_3 = 0.5$}, and \mbox{$w_4 = 0.5$}. For SegmentMeIfYouCan, we train only on Cityscapes. For BDDAnomaly, we train only on the training set of BDDAnomaly itself.



\begin{table}
  \small
  \centering
  \caption{Anomaly segmentation results on BDDAnomaly.}
  \label{tab:bdd}
  \begin{tabular}{ccc} 
    \toprule
    \textbf{Approach} & AUPR [\%] $\uparrow$ & FPR95 [\%] $\downarrow$ \\
    \midrule
    MaxSoftmax~\cite{hendrycks2017iclr} & 3.7 & 24.5  \\
    Background~\cite{blum2021ijcv} & 1.1 & 40.1 \\
    MC Dropout~\cite{gal2016icml} & 4.3 & 16.6 \\
    Confidence~\cite{devries2018arxiv} & 3.9 & 24.5 \\
    MaxLogit~\cite{hendrycks2022icml} & 5.4 & 14.0 \\
    \midrule 
    ContMAV (ours) & \textbf{96.1} & \textbf{6.9} \\
    \bottomrule
  \end{tabular}
  \vspace{-10pt}
\end{table} 

\subsection{Anomaly Segmentation}
\label{subsec:exp_ad}

The first set of experiments shows that our model achieves state-of-the-art results in anomaly segmentation, and thus also supports our first claim. Here, we aim for a binary segmentation between known classes and previously unseen classes. We report results on SegmentMeIfYouCan in~\tabref{tab:multitab}, right and BDDAnomaly in~\tabref{tab:bdd}. On SegmentMeIfYouCan, our method outperforms all baselines on FPR95 and ranks top 6 on the public leaderboard for AUPR. On the BDD datasets, our method outperforms all baselines on both metrics, providing compelling results for the task of anomaly segmentation. For the BDD datasets, in this experiment, we treat all the unknown categories as the same unknown class, without focusing on the fact they are, originally, separate classes. Our approach shows compelling results for anomaly segmentation, successfully dealing with challenging situations such as the case in which a known and an unknown object are overlapping, see~\figref{fig:results_smiyc}. While SegmentMeIfYouCan is designed specifically for anomaly segmentation, having images where the anomalous objects are prominent, the BDD dataset is more challenging since objects belonging to bicycle or motorcycle can appear in very small areas of the image (see related figures in the supplementary material), making the task of anomaly segmentation more challenging and harder to solve.  

\subsection{Open-World Semantic Segmentation}
\label{subsec:exp_ows}
The second experiment illustrates that our approach is capable of distinguishing between different unknown classes, rather than only stating whether something is known or unknown. We achieve this thanks to the feature loss function we introduced in~\eqref{eq:loss_mean_feat}. We conduct this experiment on BDDAnomaly since the test set is manually generated excluding images from the training and the validation set and thus the ground truth labels are available. Our approach is able to create multiple unknown classes, as explained in~\secref{subsec:approach_ow}. To evaluate it,for each novel class we create we report the mIoU with the ground truth category that overlaps the most to it. We report results for our method together with results we would achieve without the feature loss function. Since this task is uncommon in the literature, we report one baseline approach as a performance lower bound, that uses the background class for the unknowns and performs K-means clustering in the feature space for this class. As a performance upper bound, we report the mIoU of the three classes in closed-world settings on the original BDD100K, where there is no unknown but every class is present at training time. Results are shown in~\tabref{tab:ows}. Our approach outperforms the baseline and provides satisfying results in distinguishing among different classes. Additionally, removing the feature loss function also provides good results for open-world segmentation, outperforming the baseline by a large margin. Thus, this experiment provides support for our second claim.

\begin{table}
  \small
  \centering
  \caption{Open-world semantic segmentation results on BDDAnomaly.}
 \label{tab:ows}
 \begin{tabular}{cccc} 
   \toprule
   \multirow{3}{*}{\textbf{Approach}} & \multicolumn{3}{c}{\normalsize{\textbf{mIoU}} [\%] $\uparrow$} \\
   \cmidrule(lr){2-4}
   & Train & Motorcycle & Bicycle \\
   \midrule
   Background + cluster      & 0 & 32.3 & 32.8 \\
   ContMAV (no feat loss)   & 48.1 & 53.8 & 39.9 \\
   ContMAV (with feat loss) & \textbf{62.4} & \textbf{62.2} & \textbf{56.8} \\
   \midrule 
   Closed-world  & 72.3 & 69.3 & 60.9 \\
   \bottomrule
 \end{tabular}
\end{table}

\begin{table}
  \small
  \centering
 \caption{Class similarity results on BDDAnomaly$^*$.}
 \label{tab:cls_similarity}
   \begin{tabular}{cC{2cm}C{2cm}} 
    \toprule
    \multirow{3}{*}{\textbf{Approach}} & \multicolumn{2}{c}{\normalsize{\textbf{Accuracy}} [\%] $\uparrow$} \\
    \cmidrule(lr){2-3}
    & Motorcycle & Train  \\
    \midrule
    Baseline     & 12.5 & 9.8 \\
    ContMAV with MA   & 39.9 & 27.6 \\
    ContMAV & \textbf{58.9} & \textbf{49.9} \\
   \bottomrule
 \end{tabular}
 \vspace{-0.5cm}
\end{table}  
\subsection{Experiments on Class Similarity}
\label{subsec:exp_cs}
The third experiment shows that our approach successfully assigns to each novel class its most similar known category, supporting our third claim. For this experiment, we manually created a lookup table (see supplementary material for further details) in which each class is assigned a ground truth label indicating its most similar category. For this experiment, we used the BDDAnomaly$^*$ dataset proposed by Besnier~\etalcite{besnier2021iccv}, that is a modification of BDDAnomaly where only train and motorcycle are unknown (we report anomaly segmentation results on this dataset in the supplement). In the lookup table, the unknown class ``motorcycle'' is reported as similar to ``car'', while the unknown class ``train'' is reported as similar to ``truck''. We report one baseline that performs semantic segmentation on the known classes and has a stack of linear layers on the pre-softmax features that learns the lookup table. We compare with our same approach but taking the class that has the highest activation as most similar. We report pixel-wise accuracy results in~\tabref{tab:cls_similarity}. The results show that the classifier does not generalize well to the unknown classes. Considering only the highest activation is better than the ``specialized'' classifier, but still it is not a reliable measure of class similarity.

\begin{table}[t]
  \small
  \centering
  \caption{Ablation study on our anomaly segmentation pipeline on BDDAnomaly. $\mathcal{L}_{\mathrm{feat}}$ refers to the feature loss in~\eqref{eq:loss_mavs}, and $D_{\mathrm{cont}}$ to the contrastive decoder. ``PP'' indicates the post-processing operation used for obtaining the open-world prediction: ``Th'' for softmax thresholding, ``MA'' for maximum activation, $D_\mu$ for the minimum distance from the mean activation vector, ``Gs'' for the Gaussian inference described in~\secref{subsec:approach_ow}.}
  \begin{tabular}{cccccc}
    \toprule
    & & &  & \multicolumn{2}{c}{\textbf{BDDAnomaly}} \\
    & $\mathcal{L}_{\mathrm{feat}}$ &  $D_{\mathrm{cont}}$ & PP & AUPR [\%] $\uparrow$        & FPR95 [\%] $\downarrow$        \\
    \midrule
    A &  &  & Th & 46.9 & 93.9 \\
    B & \cmark & & Th & 76.4 & 88.6 \\
    C &  & \cmark & Th & 91.8 & 70.7 \\
    D & \cmark &  \cmark & Th & 94.1 & 54.4 \\
    \midrule
    E & \cmark & & MA & 75.9 & 89.9 \\
    F & \cmark & \cmark & MA & 93.9 & 57.6 \\
    \midrule
    G &  & \cmark & -- & 91.8 & 69.7 \\
    \midrule
    H & \cmark & & $D_\mu$  & 94.2 & 57.0 \\
    I & \cmark & \cmark & $D_\mu$ & 94.8 & 29.8 \\
    \midrule
    J & \cmark &  & Gs & 94.2 & 55.8 \\
    K & \cmark & \cmark & Gs &  \textbf{96.1} & \textbf{6.9} \\
    \bottomrule
  \end{tabular}
  \label{tab:ablation_ad}
  \vspace{-15pt}
\end{table}

\subsection{Ablation Studies}
\label{subsec:exp_abl}
Finally, we provide ablation studies to investigate the contribution of the modules we introduced. We refer to each ablation study in the tables by the letter in the first column.

\textbf{Anomaly Segmentation}. First, we perform an ablation study on the anomaly segmentation pipeline (~\tabref{tab:ablation_ad}). We investigate the contribution of the feature loss $\mathcal{L}_{\mathrm{feat}}$, of the Gaussian post-processing described in~\secref{subsec:approach_ow}, and of the contrastive decoder. We ablate different post-processing strategies. The first strategy is a softmax thresholding strategy where we consider a pixel as unknown if it has two or more activations above a threshold. The second strategy is based on the maximum softmax activation only and categorizes a pixel as unknown if its maximum activation is below a certain threshold. These two strategies yield similar performance, which is an expected outcome since they both rely on the standard final output vector. In the table, we can see that the thresholding strategy alone (A) has poor results, and its performance with the feature loss (B) is close to the performance of the maximum activation strategy with feature loss (E). Additionally, we notice how the thresholding without the feature loss but with the contrastive decoder (C) leads to better performance, that is however extremely similar to the one of the contrastive decoder only (G), suggesting that the contrastive decoder alone is better than a softmax thresholding strategy for this task. A further improvement comes from putting together the feature loss and the contrastive decoder, which leads to better results with both thresholding (D) and maximum activation (F). The other two post-processing strategies we employ are based on the output of the feature loss. One takes the minimum distance $D_\mu$ of the activation vector from the mean activation vectors we built during training, while the last one is the Gaussian querying. They lead to similar performance when the contrastive decoder is not used (H and J), and yield the top 2 performance when the contrastive decoder is used (I and K). The Gaussian querying provides a further improvement and achieves the best performance for this task.

\begin{table}[t]
  \small
  \centering
  \caption{Ablation study on our class similarity approach on BDDAnomaly$^*$. $\mathcal{L}_{\mathrm{feat}}$ refers to the feature loss in~\eqref{eq:loss_mavs}, and $D_{\mathrm{cont}}$ to the contrastive decoder. ``PP'' indicates the post-processing operation used for obtaining the open-world prediction: ``MA'' for maximum activation, $D_\mu$ for the minimum distance from the mean activation vector, ``Gs'' for the Gaussian inference described in~\secref{subsec:approach_ow}.}
  \begin{tabular}{cccccc}
    \toprule
    & & & & \multicolumn{2}{c}{\textbf{Accuracy} [\%] $\uparrow$} \\
    & $\mathcal{L}_{\mathrm{feat}}$ & $D_{\mathrm{cont}}$ & PP & Motorcycle & Train \\
    \midrule
    L & \cmark & & MA & 38.4 & 25.9 \\
    M & \cmark & \cmark & MA & 39.9 & 27.6 \\
    \midrule
    N & \cmark & & $D_\mu$ & 53.5 & 41.7 \\
    O & \cmark & \cmark & $D_\mu$ & 54.3 & 42.1 \\
    \midrule
    P & \cmark & & Gs & 57.8 & 48.6 \\
    Q & \cmark & \cmark & Gs & \textbf{58.9} & \textbf{49.9} \\
    \bottomrule
  \end{tabular}
  \label{tab:ablation_cs}
  \vspace{-10pt}
\end{table}

\textbf{Class Similarity}. The second ablation study targets the class similarity (\tabref{tab:ablation_cs}). The presence of the contrastive decoder does not substantially improve the performance, since the class similarity originates from the semantic decoder. Still, numbers when the contrastive decoder is active (M, O, Q) or inactive (L, N, P) are slightly different since the contrastive decoder affects the shared encoder via backpropagation. The performance of class similarity is poor when we rely on the standard maximum activation (L and M), while it improves when it is based on the minimum distance $D_\mu$ of the activation vector from the mean activation vectors built during training (N and O). The Gaussian post-processing achieves the best performance for both classes (P and Q), proving the effectiveness of our approach.

\section{Conclusions}
\label{sec:conclusion}

In this paper, we presented a novel approach for open-world semantic segmentation on RGB images based on a double decoder architecture. Our method manipulates the feature space of the semantic segmentation for identifying novel classes and additionally indicates the known categories that are most similar to the newly discovered ones. We implemented and evaluated our approach on different datasets and provided comparisons with other existing models and supported all claims made in this paper. The experiments suggest that our double-decoder strategy achieves compelling open-world segmentation results. In fact, with our approach, we are able to detect all anomalous regions in an image and distinguish between different novel classes.

\vspace{5pt}

\small \textbf{Acknowledgments}. This work has partially been funded by the Deutsche Forschungsgemeinschaft (DFG, German Research Foundation) under Germany’s Excellence Strategy, EXC-2070 – 390732324 – PhenoRob and by the European Union’s Horizon 2020 research and innovation programme under grant agreement No 101017008 (Harmony).
\clearpage

{
    \small
    \bibliographystyle{ieeenat_fullname}
    \bibliography{main,glorified,new}
    }
    
\clearpage
\setcounter{page}{1}
\maketitlesupplementary
The code of our approach will be available at \texttt{https://github.com/PRBonn/ContMAV}.
\appendix


\section{Further Details on Experiments}
\label{sec:sup_experiments}

\subsection{Datasets and Metrics} 
\label{subsec:data_metrics}
We use two datasets for validating our method: SegmentMeIfYouCan~\cite{chan2021neurips} and \mbox{BDDAnomaly}~\cite{hendrycks2022icml}. SegmentMeIfYouCan relies on the semantic annotations of Cityscapes~\cite{cordts2016cvpr}, and offers a public benchmark with a hidden test set for anomaly segmentation, where the goal is to segment objects that are not present on Cityscapes. Annotations are binary, since each object is either known or unknown. BDDAnomaly is a reorganization of BDD100K~\cite{yu2020cvpr}, where all images containing the classes train, motorcycle and bicycle have been discarded from the training and validation set to create an open-world test set. Since ground truth data is available for this dataset, we use it for ablation studies and experiments on class similarity. Additionally, we report results on a further modification of BDDAnomaly proposed by Besnier~\etalcite{besnier2021iccv}, which we call BDDAnomaly$^*$, where only train and motorcycle are considered as unknown classes. For metrics computation, we used the official evaluation pipeline of SegmentMeIfYouCan to enforce fairness and reproducibility\footnote{https://github.com/SegmentMeIfYouCan/road-anomaly-benchmark}. We decided to not use the area under the ROC curve (AUROC), because recently several papers showed its limitations~\cite{halligan2015er, humblot2023cvpr, wang2022nips}, as two models with the same performance may differ widely in terms of how clearly they separate in-distribution and out-of-distribution samples. In general, these works argue that AUROC is not a fair metric for comparing different approaches. This might be the reason why the official evaluation tool of SegmentMeIfYouCan, which we use in this work, does not report it.

\subsection{Experiments on Hyperparameters}
Hyperparameters search is usually a challenging problem when it comes to training neural networks. Usually, they are chosen empirically and only the configuration that works best is reported on the paper. In the following, we try to give some insight on our choice of hyperparameters and the reasoning behind them. We provide an analysis on the four hyperparameters ($\xi$, $\delta$, $\tau$, and $\eta$) in the following.

As discussed in Sec. 3.2 of the main paper, in the paragraph dedicated to the contrastive decoder, $\xi$ is the radius of the hypersphere created by the objectosphere loss~\cite{dhamija2018neurips} in Eq.~(9). In principle, this hyperparameter could take any value. However, we pair the objectosphere loss to the contrastive loss~\cite{chen2020icml} in Eq.~(8), which aims to distribute all feature vectors on the unit sphere. Thus, we expect that any choice of $\xi$ that is different from 1 would harm performance, since it would reduce the synergy between the two loss functions operating on the same decoder. We report an experiment about this in~\tabref{tab:xi}. When $\xi < 1$, the performance is not dramatically harmed because the objectosphere loss aims to make the norm of the features belonging to the known pixels greater than $\xi$. Thus, the two losses do not work against each other. In contrast, when $\xi > 1$, the two loss functions try to achieve two tasks which are incompatible (features on the unit circle and, at the same time, with norm greater than $1$), and performance suffers.
\begin{table}
    \small
   \centering
    \caption{Anomaly segmentation results on BDDAnomaly with different choices of the parameter $\xi$.}
   \label{tab:xi}
   \begin{tabular}{ccc} 
     \toprule
     \textbf{Approach} & AUPR [\%] $\uparrow$ & FPR95 [\%] $\downarrow$ \\
     \midrule
     ContMAV ($\xi = 0.75$) & 92.2 & 18.7 \\
     ContMAV ($\xi = 1.25$) & 83.4 & 55.2 \\
     \midrule
     ContMAV ($\xi = 1$) & \textbf{96.1} & \textbf{6.9} \\
     \bottomrule 
   \end{tabular}
  \end{table} 

  \begin{table}
    \small
   \centering
    \caption{Anomaly segmentation results on BDDAnomaly with different choices of the parameter $\delta$.}
   \label{tab:delta}
   \begin{tabular}{ccc} 
     \toprule
     \textbf{Approach} & AUPR [\%] $\uparrow$ & FPR95 [\%] $\downarrow$ \\
     \midrule
     ContMAV ($\delta = 0.4$) & 86.6 & 41.2 \\
     ContMAV ($\delta = 0.8$) & 89.1 & 30.1 \\
     \midrule
     ContMAV ($\delta = 0.6$) & \textbf{96.1} & \textbf{6.9} \\
     \bottomrule 
   \end{tabular}
  \end{table} 
 
The threshold $\delta$, which we also introduced in Sec. 3.2, in the paragraph dedicated to the post-processing, is our ``unknown-ness threshold''. In fact, we obtain a score \mbox{$s_{\mathrm{unk}} \in [0, 1]$} and have to decide whether a pixel belongs to an unknown category based on this score. The score is given by
\begin{equation}
    s_{\mathrm{unk}} = \frac{1}{2} \, \Big( s_{\mathrm{unk}}^{\mathrm{sem}} + s_{\mathrm{unk}}^{\mathrm{cont}} \Big),
    \label{eq:score_final}
  \end{equation}
where $s_{\mathrm{unk}}^{\mathrm{seg}}$ and $s_{\mathrm{unk}}^{\mathrm{cont}}$ are the scores coming from the semantic and the contrastive decoders, respectively. Notice that, since the final score is a standard mean of the two, setting the threshold to a low value would make us label a pixel as unknown also in the case in which only one score is high but the other is not. This would create a lot of false positives, and we expect performance aligned with models G and J in Tab.~5 of the main paper. Those two models, in fact, only have one active decoder, and setting a low $\delta$ causes a similar behavior. Setting the threshold too high is, in contrast, achievable only when both decoder heads are very confident in their prediction of unknown, and it could cause a high number of false negatives. Thus, we choose $\delta = 0.6$, that is a good compromise and provides good results (see~\tabref{tab:delta}).

We do not optimize the temperature parameter of the contrastive loss $\tau$ and perform all experiments with $\tau = 0.1$, as suggested by Chen~\etalcite{chen2020icml}.

The hyperparameter $\eta$, also introduced in Sec. 3.2, in the paragraph dedicated to the post-processing, does not affect the prediction of a pixel as unknown, but it plays a role in the class discovery. In fact, it represents the minimum distance needed to decide whether a feature categorized as unknown is a class of its own and does not belong to any of the already-discovered new classes. Setting this threshold heavily depends on the data distribution. A very high threshold would create a lot of classes, and its usefulness would be limited. On the other hand, a low threshold would put all classes together, providing nothing more than an anomaly segmentation. We report results in~\tabref{tab:eta}, where we also report the number $N_U$ of new classes created, for which the ground truth value is $3$ (i.e., the number of unknown classes in BDDAnomaly).

\begin{table}
    \small
    \centering
    \caption{Class discovery results on BDDAnomaly with different choices of the parameter $\eta$. For each class of interest, the discovered one with greater IoU is chosen and reported.}
   \label{tab:eta}
   \begin{tabular}{ccccc} 
     \toprule
     \multirow{2}{*}{\textbf{Approach}} & \multicolumn{3}{c}{\normalsize{\textbf{mIoU}} [\%] $\uparrow$} & \multirow{2}{*}{$N_U$} \\
     & Train & Motorcycle & Bicycle & \\
     \midrule
     ContMAV ($\eta = 0.3$)      & 0.0 & 23.4 & 0.0 & 1\\
     ContMAV ($\eta = 0.9$)   & 30.5 & 31.1 & 18.9 & 12\\
     \midrule
     ContMAV ($\eta = 0.6$) & \textbf{62.4} & \textbf{62.2} & \textbf{56.8} & 4\\
     \bottomrule
   \end{tabular}
  \end{table}

\subsection{Further Details on Anomaly Segmentation}
In Sec. 4 of the main paper, we reported extensive experiments on SegmentMeIfYouCan~\cite{chan2021neurips} and BDDAnomaly~\cite{hendrycks2022icml}. SegmentMeIfYouCan is a public benchmark for anomaly segmentation, with a hidden test set and a public leaderboard. Our method, called ContMAV, ranks first overall on three out of five metrics, namely FPR95, PPV and mean F1, and it ranks fourth on AUPR and sixth on sIoU. Further details and baselines results can be found on SegmentMeIfYouCan's official website: \small{\texttt{https://segmentmeifyoucan.com/leaderboard}}.
\normalsize
Besnier~\etalcite{besnier2021iccv} proposed a modification of BDDAnomaly~\cite{hendrycks2022icml}, where only two classes (train and motorcycle) are considered as unknown. We call this dataset BDDAnomaly$^*$. Differently from BDDAnomaly, the images containing bicycle are not discarded from the training and validation set, but are kept and bicycle is considered a known class. We test our method also on this dataset, using the same training details and parameters discussed above. We report our results in~\tabref{tab:bddstar}.

\subsection{Further Details on Class Similarity}
In Sec. 4.3 of the main paper, we reported our experiment on class similarity, and mentioned the creation of a lookup table in which each class is assigned a ground truth label indicating its most similar category. We chose the most similar class based on the relevance in an autonomous driving scenario. For example, truck is paired to bus since one could expect a similar behavior between these two traffic participants. Some classes, such as sky, are not assigned any label for the most similar category. The lookup table is reported in~\tabref{tab:lookup}. For this experiment, we decided to use BDDAnomaly$^*$ because we did not find a valid correspondence for the class bicycle. The only vehicle that belongs to known classes is car, and in fact our method on BDDAnomaly achieves, for bicycle, a 43.2$\%$ similarity score with car. A more modern dataset, with more vehicle classes such as electric scooters, would provide better candidates for class similarity. 

\subsection{Architectural Choices}
As reported in~\secref{subsec:arch}, we used a modified version of ResNet34. Still, our contribution does not include any of the modules presented there, such as the NonBottleneck-1D block or the average pyramid pooling module, whose contributions are reported in the related papers~\cite{romera2018tits, zhao2017cvpr-pspn}. Therefore, we do not provide ablation studies on these components, but rather all of our models and ablations use them. 

\begin{table}
    \small
   \centering
    \caption{Anomaly segmentation results on BDDAnomaly*.}
   \label{tab:bddstar}
   \begin{tabular}{ccc} 
     \toprule
     \textbf{Approach} & AUPR [\%] $\uparrow$ & FPR95 [\%] $\downarrow$ \\
     \midrule
     MaxSoftmax~\cite{hendrycks2017iclr} & 80.1 & 63.5 \\
     Background~\cite{blum2021ijcv} & 75.3 & 68.1 \\
     MC Dropout~\cite{gal2016icml} & 82.6 & 61.1 \\
     ODIN~\cite{liang2018iclr} & 81.7 & 60.6 \\
     ObsNet + LAA~\cite{besnier2021iccv} & 82.8 & 60.3 \\
     \midrule 
     ContMAV (ours) & \textbf{92.9} & \textbf{43.9} \\
     \bottomrule 
   \end{tabular}
  \end{table} 

  \begin{table}
    \small
   \centering
    \caption{Look-up table for class similarity. The unknowns are specified in the context of BDDAnomaly$^*$.}
   \label{tab:lookup}
   \begin{tabular}{ccc} 
     \toprule
     \textbf{Category} & \textbf{Most Similar} & \textbf{Type} \\
     \midrule
     Road & Sidewalk & stuff \\
     Sidewalk & Road & stuff \\
     Building & Wall & stuff \\
     Wall & Fence & stuff \\
     Fence & Wall & stuff \\
     Pole & Sign & stuff \\
     Light & Sign & stuff \\
     Vegetation & Terrain & stuff \\
     Terrain & Vegetation & stuff \\
     Sky & -- & stuff \\
     Person & Rider & thing \\
     Rider & Person & thing \\
     Car & Truck & thing \\
     Truck & Bus & thing \\
     Bus & Truck & thing \\
     Bicycle & -- & thing \\
     \midrule
     Train & Truck & thing, unknown \\
     Motorcycle & Car & thing, unknown \\
     \bottomrule 
   \end{tabular}
  \end{table}

\section{Further Details on the Contrastive Decoder}
\label{sec:sup_contrastive}
The contrastive decoder, which we explain in details in~\secref{subsec:approach_ow}, is optimized with a combination of two loss functions, namely the objectosphere and the contrastive loss. \figref{fig:contrastive_dec} intuitively shows the idea behind it, and what the ideal output in the 2D case would be. However, the feature vectors that the contrastive decoder predicts are $K$-dimensional, where $K$ is the number of known classes (\ie, 19 in our case). In order to verify whether the output of the decoder is aligned with our expectation, we define two thresholds $\zeta$ and $\rho$. Then, given $\b{f}_p^d$, \ie, the feature predicted at pixel $p$ from the contrastive decoder, we want ${1 - \zeta < ||\b{f}_p^d||_2 < 1 + \zeta}$ for all $\b{f}_p^d$ whose ground truth label is a known class, and $||\b{f}_p^d||_2 < \rho$ for all $\b{f}_p^d$ whose ground truth label is an unknown class. The former means that the norms of the vectors belonging to known classes should be in a ``tube'' of radius $\zeta$ around $1$, which is our $\xi$ parameter as explained in~\tabref{tab:xi}. The latter means that the norms of the vectors belonging to unknown classes (which, at training time, are the unlabeled portions of the image), should be smaller than $\rho$. We choose $\zeta = 0.2$ and $\rho = 0.4$, and we find that $86.5 \%$ of the vectors belonging to known classes fall into the tube, and that $79.9 \%$ of the vectors belonging to unknown classes are smaller than $\rho$. This verifies that the output is aligned to our expectation. To visually show the result, we would need to apply a dimensionality reduction approach such as principal component analysis. However, linear dimensionality reduction techniques always lead to loss of information, and the new dimensions may offer no concrete interpretability.

\begin{table}
  \small
  \centering
  \caption{Architectural Efficiency}
  \label{tab:efficiency}
  \begin{tabular}{ccc} 
    \toprule
    \textbf{Approach} & GFLOPs $\downarrow$ & Training Parameters $\downarrow$ \\
    \midrule
    Maskomaly~\cite{hendrycks2017iclr} & 937 & 215M  \\
    Mask2Anomaly~\cite{blum2021ijcv} & 258 & \textbf{23M} \\
    \midrule 
    ContMAV (ours) & \textbf{84} & 48M \\
    \bottomrule
  \end{tabular}
  \vspace{-10pt}
\end{table} 

\section{Architectural Efficiency}
\label{sec:sup_efficiency}
As pointed out in~\secref{subsec:arch}, we designed our neural network in order to be lightweight and faster at inference time. The architecture design choices explained in~\secref{subsec:arch} allow inference on an image at $10$ Hz. Additionally, we report the number of parameters and the GFLOPs of our model together with two state-of-the-art models from the SegmentMeIfYouCan public benchmark with code available in~\tabref{tab:efficiency}. We show that our architecture is competitive and performs very well in terms of efficiency. 

\section{Qualitative Results}
\label{sec:sup_qualitative}

\begin{figure*}[hp]
    \centering
    \includegraphics[width=0.99\linewidth]{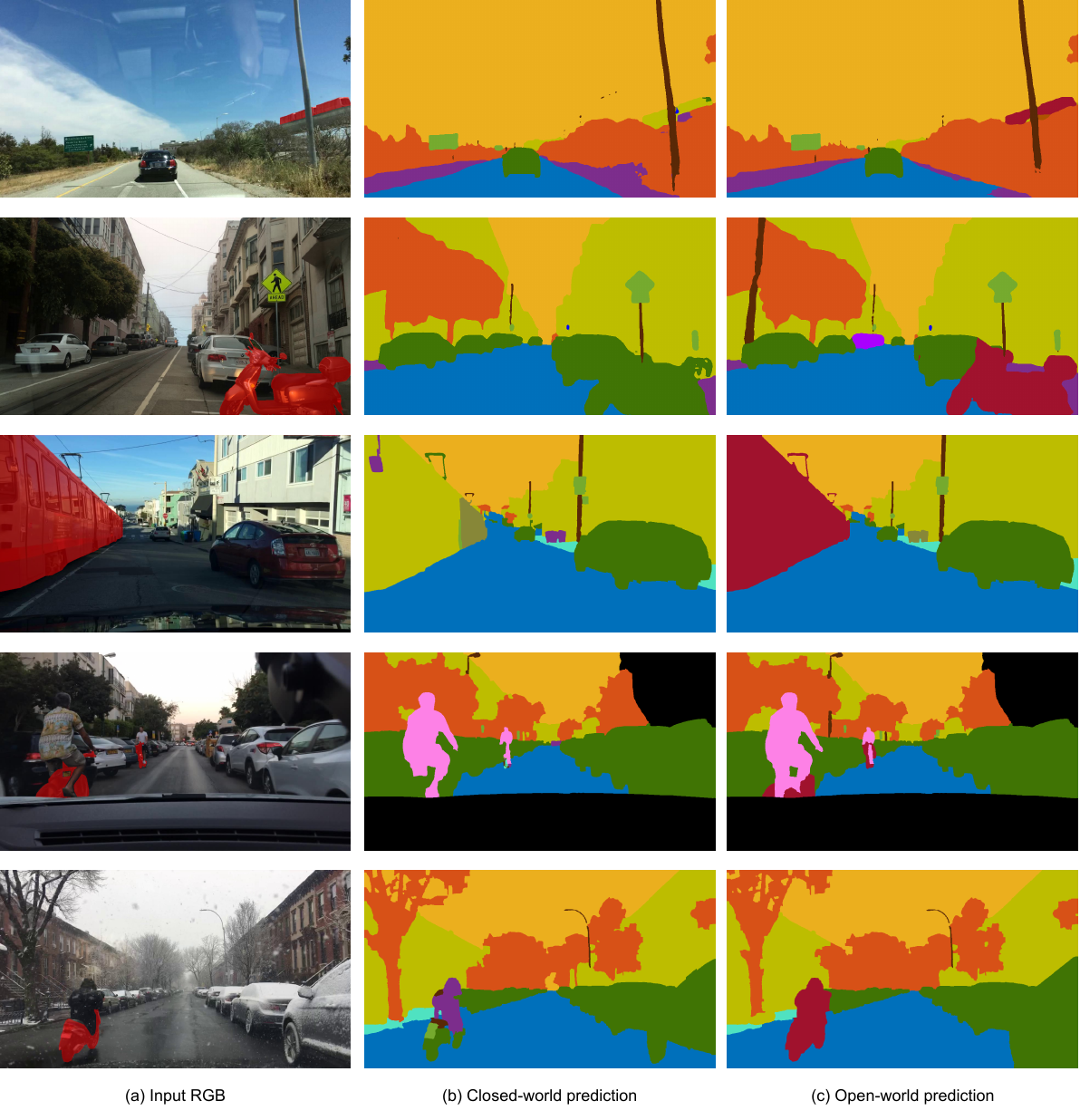}
    \caption{Anomaly segmentation results from the test set of BDDAnomaly. We show the input RGB overlayed with the ground truth unknown mask (a), the prediction of our closed-world model (b), and the prediction of our approach for open-world segmentation (c). In the open-world prediction, the unknown class is shown in red. Notice how the two models, that are both trained on BDDAnomaly, perform similarly on known classes, demonstrating that our approach does not degrade closed-world performance.}
    \label{fig:results_bdd}
  \end{figure*}

\begin{figure*}[hp]
    \centering
    \includegraphics[width=0.9\linewidth]{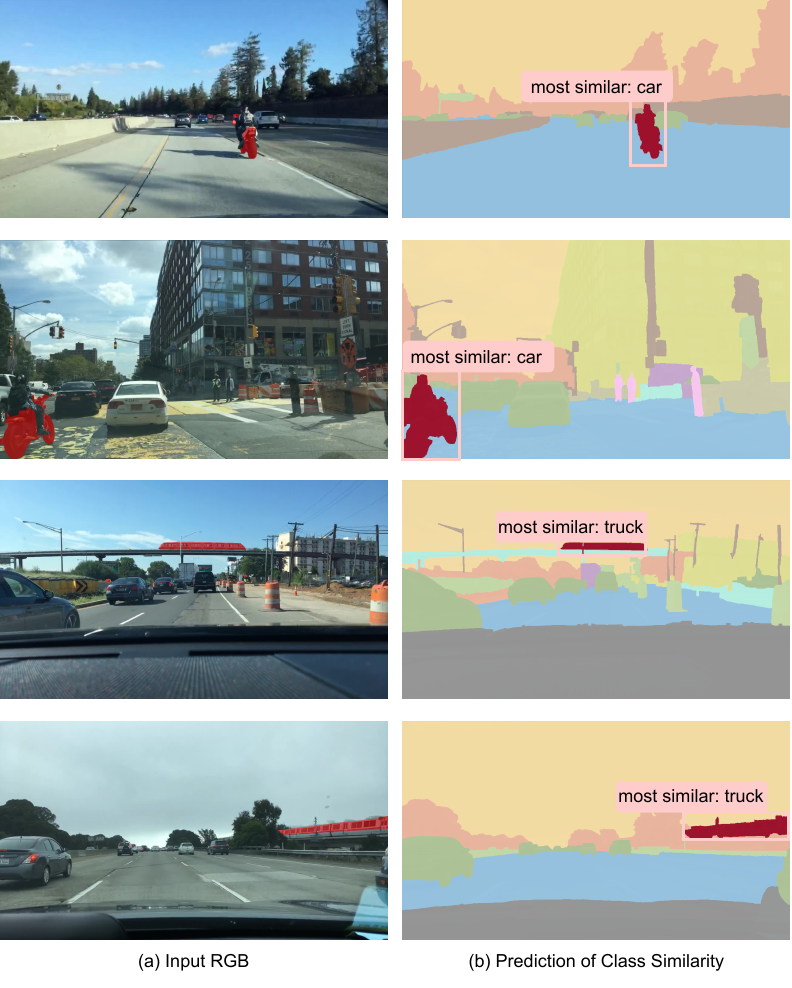}
    \caption{Class similarity results from the test set of BDDAnomaly. We show the input RGB overlayed with the ground truth unknown mask (a) and the prediction of our class similarity pipeline (b). In the open-world prediction, the unknown class is shown in red, and the overall semantic segmentation is shown in transparency.}
    \label{fig:results_bdd_cs}
  \end{figure*}

We provide further qualitative results on the validation set of SegmentMeIfYouCan and the test set of BDDAnomaly in~\figref{fig:results_smiyc} and~\figref{fig:results_bdd}, respectively. Additionally, we report qualitative results on the test set of BDDAnomaly for class similarity in~\figref{fig:results_bdd_cs}.

\section{Limitations and Future Works}
\label{sec:sup_limitations}
As shown in the various experiments, our approach achieves state-of-the-art results on different datasets on both, anomaly segmentation and novel class discovery. Still, our approach presents some limitations which offer interesting avenues for future work in order to make the approach more robust and performing. In particular, the semantic decoder builds a mean activation vector, or average class prototype, for each class and the dimension of this descriptor is equal to the number of known classes. When not many classes are available at training time, this descriptor collapses to a few dimensions, which might be not descriptive enough for ensuring a reliable novel class discovery where many new classes can be found. The contrastive decoder instead leverages the unlabeled portions of the image as unknowns available at training time to train the objectosphere loss (basically following the concept of ``known unknowns'' introuced by Bendale~\etalcite{bendale2016cvpr}), and would suffer from a fully labeled dataset where no pixel is left with no ground truth annotation. Additionally, we provide open-world semantic segmentation (\ie, anomaly segmentation and novel class discovery) only, but no instance are segmented. An interesting research avenue is to extend this work in the direction of open-world panoptic segmentation.

\begin{figure*}[hp!]
  \centering
  \includegraphics[width=0.99\linewidth]{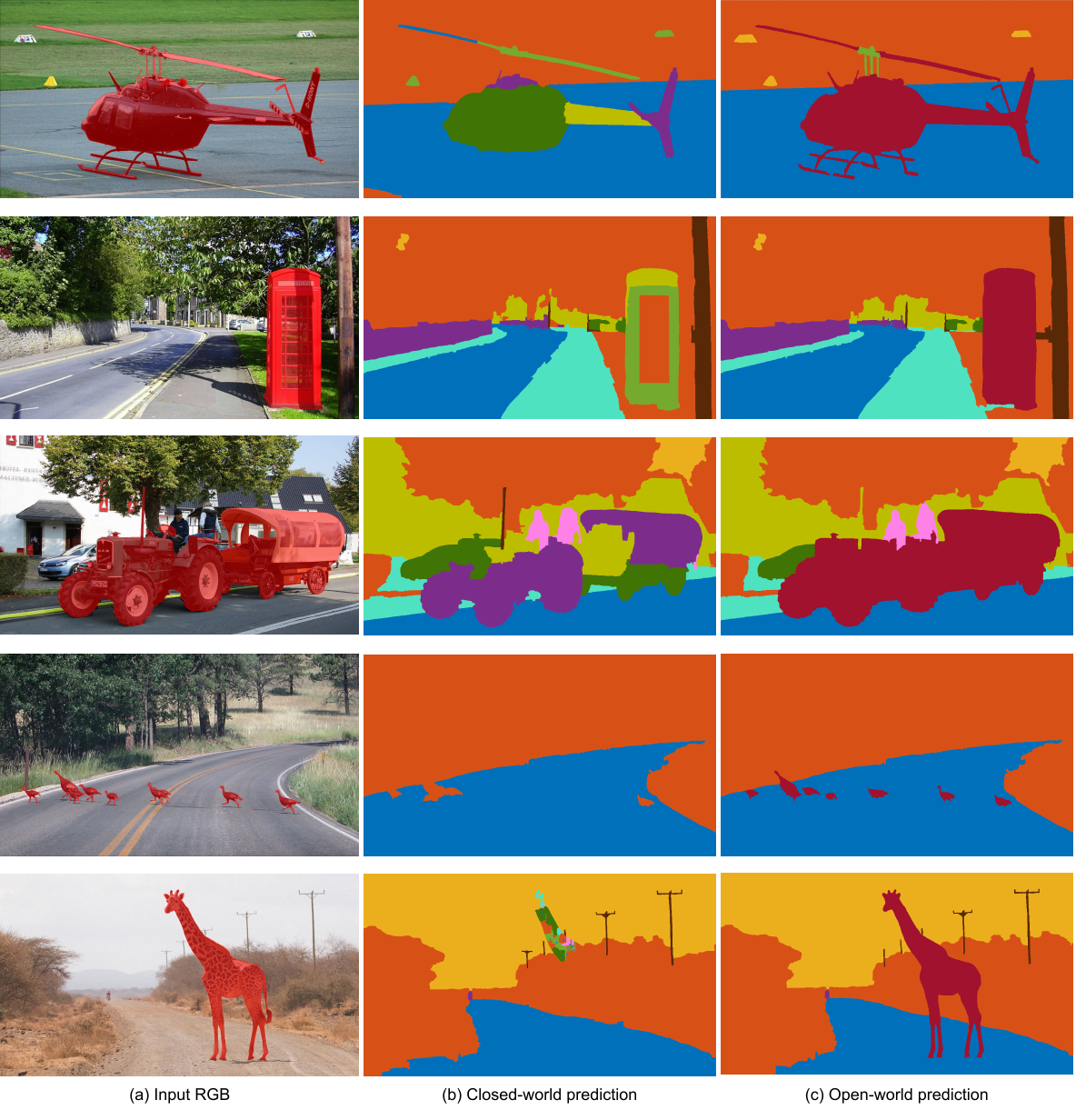}
  \caption{Anomaly segmentation results from the validation set of SegmentMeIfYouCan. We show the input RGB overlayed with the ground truth unknown mask (a), the prediction of our closed-world model (b), and the prediction of our approach for open-world segmentation (c). In the open-world prediction, the unknown class is shown in red. Notice how the two models, that are both trained on CityScapes, perform similarly on known classes, demonstrating that our approach does not degrade closed-world performance.}
  \label{fig:results_smiyc}
\end{figure*}

  \begin{figure*}[hp]
    \centering
    \includegraphics[width=0.95\linewidth]{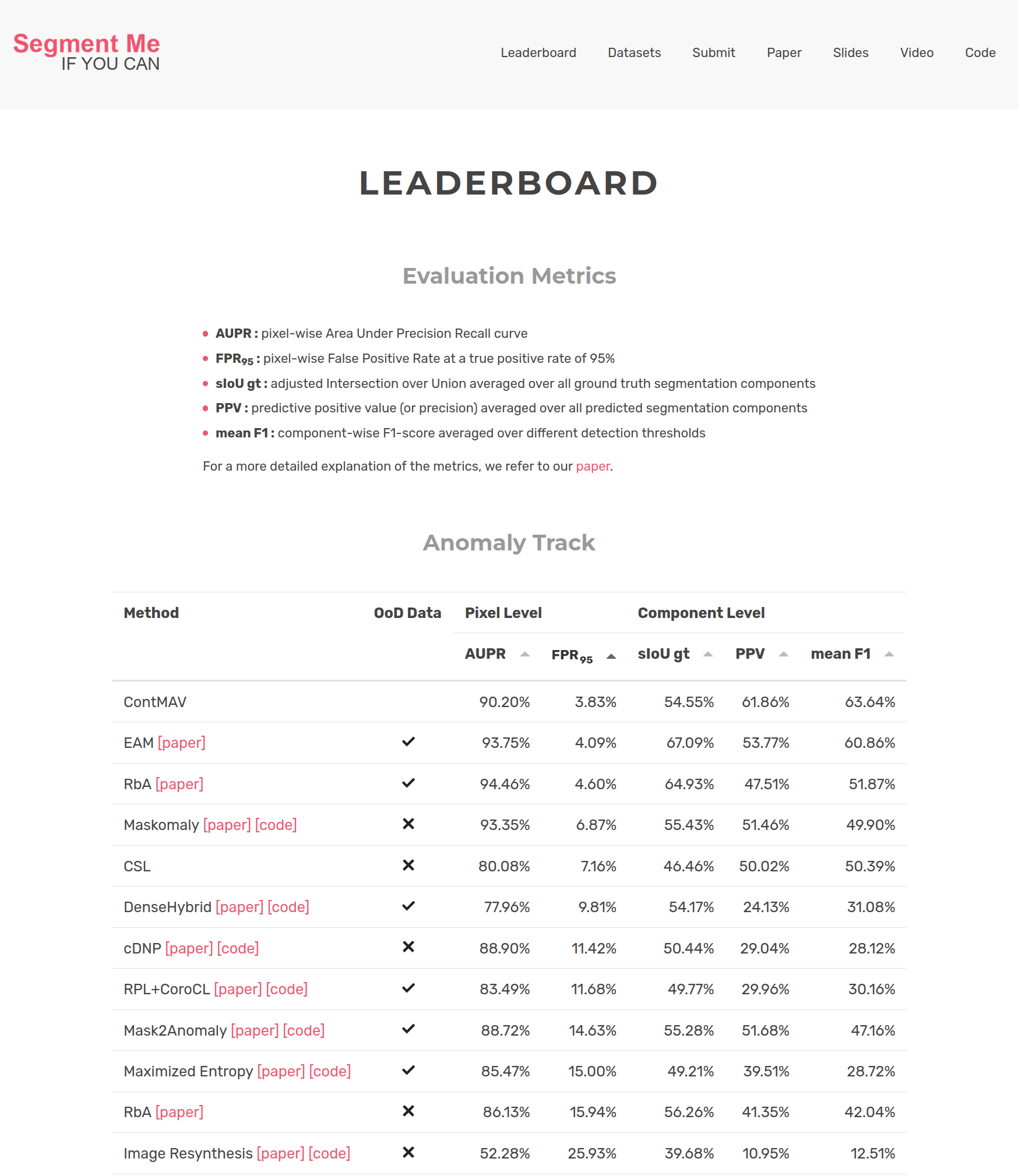}
    \caption{Screenshot of the top methods in the public leaderboard of SegmentMeIfYouCan, taken on November 21st 2023. Our method, ContMAV, is the top approach for FPR95, PPV, and mean F1. To preserve anonymity, paper and code will be attached to the benchmark submission upon acceptance.}
  \end{figure*}


\end{document}